\documentclass[10pt,twocolumn,letterpaper]{article}

\usepackage{cvpr}
\usepackage{times}
\usepackage{epsfig}
\usepackage{graphicx}
\usepackage{amsmath}
\usepackage{amssymb}
\usepackage{multirow}
\usepackage{tabularx}
\usepackage{algpseudocode}
\usepackage[ruled,vlined,linesnumbered]{algorithm2e}
\usepackage{color,soul}
\usepackage{gensymb}
\usepackage{slashbox}
\usepackage{cite}

\def\eqnvspace{{\vspace{-0.5mm}}}
\def\textvspace{{\vspace{-3mm}}}
\def\figvspace{{\vspace{-3mm}}}
\newcommand{\Paragraph}[1]{\vspace{-0mm} \noindent \textbf{#1} \hspace{0mm}}
\newcommand{\Section}[1]{\vspace{-1mm} \section{#1} \vspace{-1mm}}
\newcommand{\SubSection}[1]{\vspace{-0mm} \subsection{#1} \vspace{-1mm}}
\newcommand{\SubSubSection}[1]{\vspace{-1mm} \subsubsection{#1} \vspace{-1mm}}

\DeclareMathOperator*{\argmin}{arg\,min}

\usepackage[pagebackref=true,breaklinks=true,letterpaper=true,colorlinks,bookmarks=false]{hyperref}

 \cvprfinalcopy 

\def\cvprPaperID{615} 

\ifcvprfinal\fi
\begin{document}

\title{Learning Deep Models for Face Anti-Spoofing: Binary or Auxiliary Supervision}

\author{ Yaojie Liu\thanks{denotes equal contribution by the authors.} $\quad$ Amin Jourabloo\footnotemark[1] $\quad$ Xiaoming Liu \\
Department of Computer Science and Engineering\\
Michigan State University, East Lansing MI 48824\\
{\tt\small \{liuyaoj1,jourablo,liuxm\}@msu.edu }}

\maketitle

\begin{abstract}
 
Face anti-spoofing is crucial  to prevent face recognition systems from a security breach. 
Previous deep learning approaches formulate face anti-spoofing as a binary classification problem. 
Many of them struggle to grasp adequate spoofing cues and generalize poorly. 
In this paper, we argue the importance of auxiliary supervision to guide the learning toward discriminative and generalizable cues. 
A CNN-RNN model is learned to estimate the face depth with pixel-wise supervision, and to estimate rPPG signals with sequence-wise supervision. 
The estimated depth and rPPG are fused to distinguish live vs.~spoof faces. 
Further, we introduce a new face anti-spoofing database that covers a large range of illumination, subject, and pose variations. 
Experiments show that our model achieves the state-of-the-art results on both intra- and cross-database testing. 
   
\end{abstract}

\Section{Introduction}
With applications in phone unlock, access control, and security, biometric systems are widely used in our daily lives, and face is one of the most popular biometric modalities. 
While face recognition systems gain popularity, attackers present face spoofs (i.e., presentation attacks, PA) to the system and attempt to be authenticated as the genuine user. 
The face PA include printing a face on paper (print attack), replaying a face video on a digital device (replay attack), wearing a mask (mask attack), etc.
To counteract PA, face anti-spoofing techniques~\cite{frischholz2000biold,chetty2006multi,frischholz2003avoiding,li2004live} are developed to detect PA {\it prior to} a face image being recognized.
Therefore, face anti-spoofing is vital to ensure that face recognition systems are robust to PA and safe to use.

\begin{figure}[t!]
	\centering
	\small
	\includegraphics[width=0.92\linewidth]{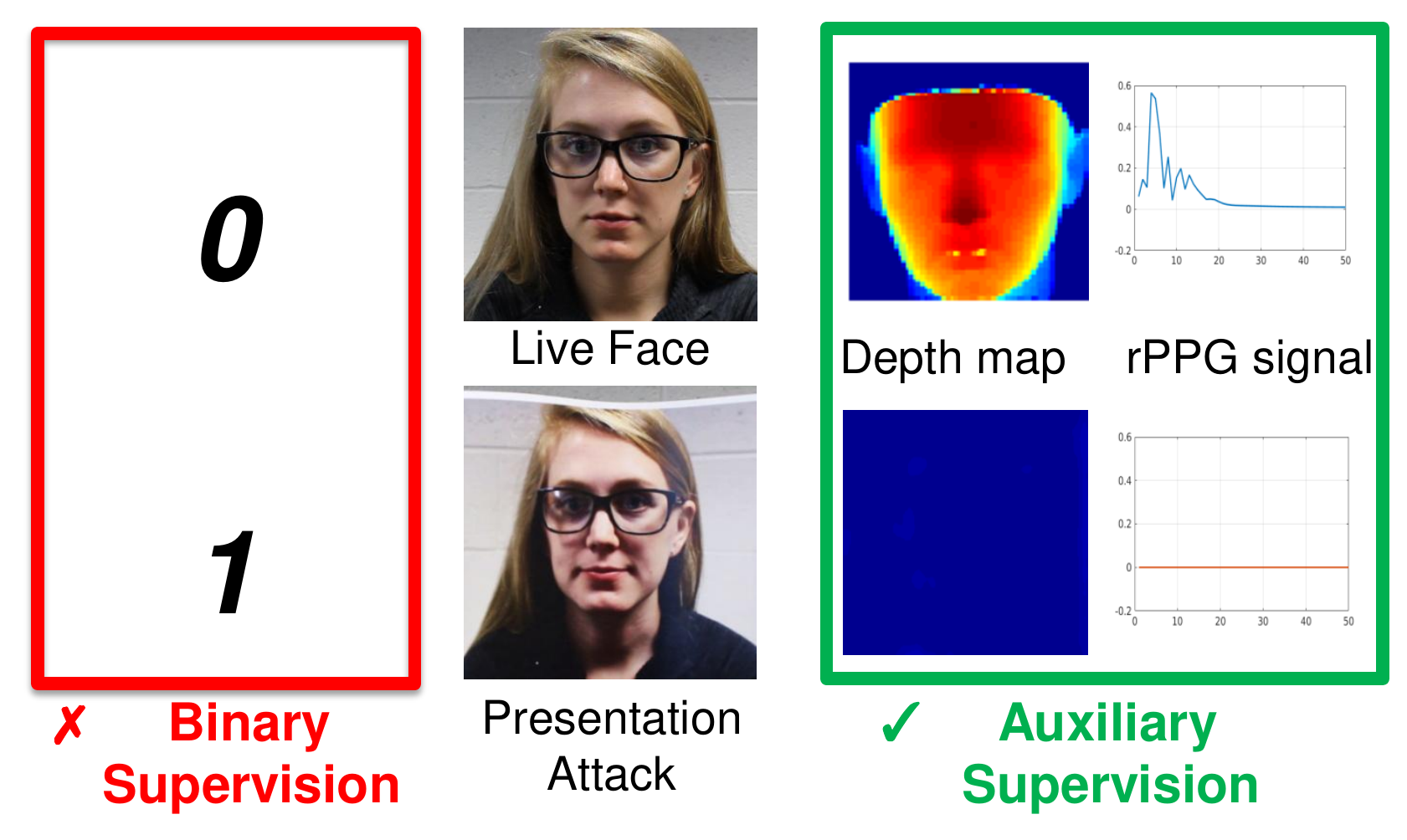} 
	\caption{\small Conventional CNN-based face anti-spoof approaches utilize the binary supervision, which may lead to overfitting given the enormous solution space of CNN. This work designs a novel network architecture to leverage two auxiliary information as supervision: the depth map and rPPG signal, with the goals of improved generalization and explainable decisions during inference.}
	\label{fig:1}
	\figvspace
\end{figure}

RGB image and video are the standard input to face anti-spoofing systems, similar to face recognition systems. 
Researchers start the texture-based anti-spoofing approaches by feeding handcrafted features to binary classifiers~\cite{maatta2011face,de2012lbp,de2013can,mirjalili2017soft,komulainen2013context,yang2013face,patel2016secure,boulkenafet2017face}.
Later in the deep learning era, several Convolutional Neural Networks (CNN) approaches utilize softmax loss as the supervision~\cite{li2016original,patel2016cross,yang2014learn,feng2016integration}.
It appears almost all prior work regard the face anti-spoofing problem as merely a $\textit{binary}$ (live vs.~spoof) classification problem.

There are two main issues in learning deep anti-spoofing models with binary supervision. 
First, there are different levels of image degradation, namely $\textit{spoof patterns}$, comparing a spoof face to a live one, which consist of skin detail loss, color distortion, moir\'e pattern, shape deformation and spoof artifacts (e.g., reflection)~\cite{li2004live,patel2016secure}.
A CNN with softmax loss might discover \textit{arbitrary} cues that are able to separate the two classes, such as screen bezel, but not the \textit{faithful} spoof patterns.
When those cues disappear during testing, these models would fail to distinguish spoof vs.~live faces and result in poor generalization.
Second, during the testing, models learnt with binary supervision will only generate a binary decision without \textit{explanation} or \textit{rationale} for the decision.
In the pursuit of Explainable Artificial Intelligence~\cite{darpa-xai}, it is desirable for the learnt model to generate the spoof patterns that support the final binary decision. 

To address these issues, as shown in Fig.~\ref{fig:1}, we propose a deep model that uses the supervision from both the {\it spatial} and {\it temporal auxiliary information} rather than binary supervision, for the purpose of robustly detecting face PA from a face video.
These auxiliary information are acquired based on our domain knowledge about the key {\it differences} between live and spoof faces, which include two perspectives: spatial and temporal. 
From the spatial perspective, it is known that live faces have face-like depth, e.g., the nose is closer to the camera than the cheek in frontal-view faces, while faces in print or replay attacks have flat or planar depth, e.g., all pixels on the image of a paper have the same depth to the camera.
Hence, depth can be utilized as auxiliary information to supervise both live and spoof faces. 
From the temporal perspective, it was shown that the normal rPPG signals (i.e., heart pulse signal) are detectable from live, but not spoof, face videos~\cite{liu20163d2,nowara2017ppgsecure}.
Therefore, we provide different rPPG signals as auxiliary supervision, which guides the network to learn from live or spoof face videos respectively.
To enable both supervisions, we design a network architecture with a short-cut connection to capture different scales and a novel non-rigid registration layer to handle the motion and pose change for rPPG estimation.

Furthermore, similar to many vision problems, data plays a significant role in training the anti-spoofing models. 
As we know, camera/screen quality is a critical factor to the quality of spoof faces. 
Existing face anti-spoofing databases, such as NUAA~\cite{tan2010face}, CASIA~\cite{zhang2012face}, Replay-Attack~\cite{Chingovska_BIOSIG-2012}, and MSU-MFSD~\cite{WenTIFS15}, were collected $3-5$ years ago. 
Given the fast advance of consumer electronics, the types of equipment (e.g., cameras and spoofing mediums) used in those data collection are outdated compared to the ones nowadays, regarding the resolution and imaging quality. 
More recent MSU-USSA~\cite{patel2016secure} and OULU databases~\cite{OULU_NPU_2017} have subjects with fewer variations in poses, illuminations, expressions (PIE). 
The lack of necessary variations would make it hard to learn an effective model. 
Given the clear need for more advanced databases, we collect a face anti-spoofing database, named Spoof in the Wild Database (SiW). 
SiW database consists of $165$ subjects, $6$ spoofing mediums, and $4$ sessions covering variations such as PIE, distance-to-camera, etc. 
SiW covers much larger variations than previous databases, as detailed in Tab.~\ref{tab:dataset} and Sec.~\ref{sec:database}.
The main contributions of this work include:

$\diamond$  We propose to leverage novel auxiliary information (i.e., depth map and rPPG) to supervise the CNN learning for improved generalization.

$\diamond$  We propose a novel CNN-RNN architecture for end-to-end learning the depth map and rPPG signal.

$\diamond$  We release a new database that contains variations of PIE, and other practical factors. We achieve the state-of-the-art performance for face anti-spoofing.

\Section{Prior Work}

We review the prior face anti-spoofing works in three groups: texture-based methods, temporal-based methods,  and remote photoplethysmography methods. 

\Paragraph{Texture-based Methods} Since most face recognition systems adopt only RGB cameras, using texture information has been a natural approach to tackling face anti-spoofing.
Many prior works utilize hand-crafted features, such as LBP~\cite{maatta2011face,de2012lbp,de2013can}, HoG~\cite{komulainen2013context,yang2013face}, SIFT~\cite{patel2016secure} and SURF~\cite{boulkenafet2017face}, and adopt traditional classifiers such as SVM and LDA. 
To overcome the influence of illumination variation, they seek solutions in a different input domain, such as HSV and YCbCr color space~\cite{boulkenafet2015face,boulkenafet2016face}, and Fourier spectrum~\cite{li2004live}.   

As deep learning has proven to be effective in many computer vision problems, there are many recent attempts of using CNN-based features or CNNs in face anti-spoofing~\cite{li2016original,patel2016cross,yang2014learn,feng2016integration}. 
Most of the work treats face anti-spoofing as a simple {\it binary} classification problem by applying the softmax loss.
For example, \cite{li2016original,patel2016cross} use CNN as feature extractor and fine-tune from ImageNet-pretrained CaffeNet and VGG-face. 
The work of~\cite{li2016original,feng2016integration} feed different designs of the face images into CNN, such as multi-scale faces and hand-crafted features, and directly classify live vs.~spoof.
One prior work that shares the similarity with ours is~\cite{face-anti-spoofing-using-patch-and-depth-based-cnns}, where Atoum \etal~propose a two-steam CNN-based anti-spoofing method using texture and depth. 
We advance~\cite{face-anti-spoofing-using-patch-and-depth-based-cnns} in a number of aspects, including fusion with temporal supervision (i.e., rPPG), finer architecture design, novel non-rigid registration layer, and comprehensive experimental support. 

\Paragraph{Temporal-based Methods} 
One of the earliest solutions for face anti-spoofing is based on temporal cues such as eye-blinking~\cite{patel2016cross,pan2007eyeblink}. 
Methods such as~\cite{deep-convolutional-dynamic-texture-learning,kollreider2007real} track the motion of mouth and lip to detect the face liveness. 
While these methods are effective to typical paper attacks, they become vulnerable when attackers present a replay attack or a paper attack with eye/mouth portion being cut. 

There are also methods relying on more general temporal features, instead of the specific facial motion. 
The most common approach is frame concatenation. 
Many handcrafted feature-based methods may improve intra-database testing performance by simply concatenating the features of consecutive frames to train the classifiers~\cite{komulainen2013complementary,de2012lbp,boulkenafet2015face}. 
Additionally, there are some works proposing temporal-specific features, e.g., Haralick features~\cite{agarwal2016face}, motion mag~\cite{bharadwaj2014face}, and optical flow~\cite{bao2009liveness}.
In the deep learning era, Feng~\etal feed the optical flow map and Shearlet image feature to CNN~\cite{feng2016integration}. 
In~\cite{xu2015learning}, Xu~\etal propose an LSTM-CNN architecture to utilize temporal information for binary classification. 
Overall, all prior methods still regard face anti-spoofing as a binary classification problem, and thus they have a hard time to generalize well in the cross-database testing.
In this work, we extract discriminative temporal information by learning the rPPG signal of the face video. 

\begin{figure}[t!]
	\centering
	\small
	\includegraphics[width=\linewidth]{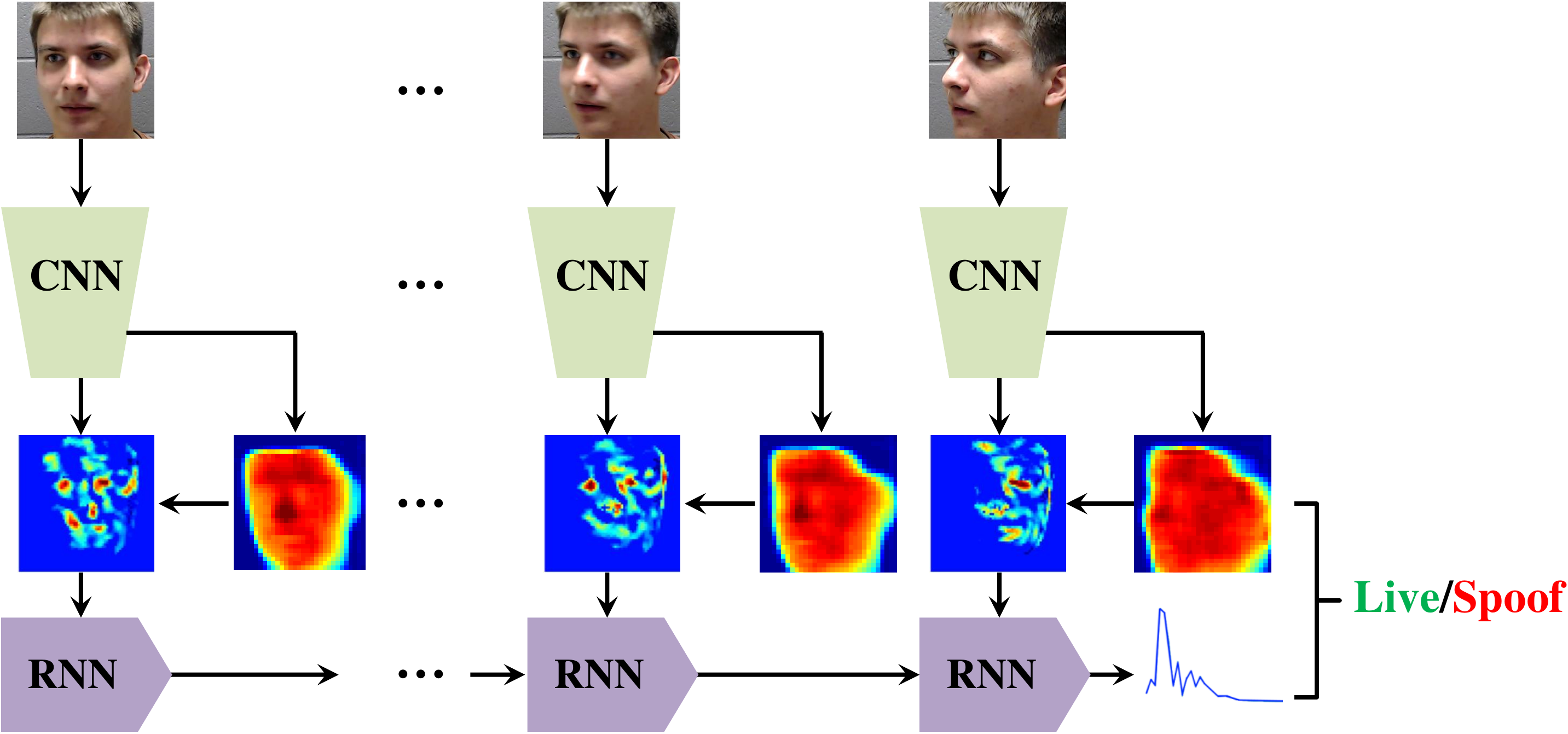} 
	\caption{\small The overview of the proposed method.}
	\label{fig:overview}
	\figvspace
\end{figure}

\Paragraph{Remote Photoplethysmography (rPPG)} 
Remote photoplethysmography (rPPG) is the technique to track vital signals, such as heart rate, without any contact with human skin~\cite{de2013robust,po2017block,bobbia2016remote,tulyakov2016self, wu2016motion}. 
Research starts with face videos with no motion or illumination change to videos with multiple variations. 
In~\cite{de2013robust}, Haan~\etal  estimate rPPG signals from RGB face videos with lighting and motion changes. 
It utilizes color difference to eliminate the specular reflection and estimate two orthogonal chrominance signals. 
After applying the Band Pass Filter (BPM), the ratio of the chrominance signals are used to compute the rPPG signal.




rPPG has previously been utilized to tackle face anti-spoofing~\cite{liu20163d2,nowara2017ppgsecure}. 
In ~\cite{liu20163d2}, rPPG signals are used for detecting the $3$D mask attack, where the live faces exhibit a pulse of heart rate unlike the $3$D masks. 
They use rPPG signals extracted by~\cite{de2013robust} and compute the correlation features for classification.  
Similarly, Magdalena \etal~\cite{nowara2017ppgsecure} extract rPPG signals (also via~\cite{de2013robust}) from three face regions and two non-face regions, for detecting print and replay attacks. 
Although in replay attacks, the rPPG extractor might still capture the normal pulse, the combination of multiple regions can differentiate live vs.~spoof faces. 
While the analytic solution to rPPG extraction~\cite{de2013robust} is easy to implement, we observe that it is sensitive to PIE variations. 
Hence, we employ a novel CNN-RNN architecture to {\it learn} a mapping from a face video to the rPPG signal, which is not only robust to PIE variations, but also discriminative to live vs.~spoof.


\Section{Face Anti-Spoofing with Deep Network}
The main idea of the proposed approach is to guide the deep network to focus on the {\it known spoof patterns} across spatial and temporal domains, rather than to extract any cues that could separate two classes but are not generalizable.
As shown in Fig.~\ref{fig:overview}, the proposed network combines CNN and RNN architectures in a coherent way. 
The CNN part utilizes the depth map supervision to discover subtle texture property that leads to distinct depths for live and spoof faces. 
Then, it feeds the estimated depth and the feature maps to a novel \textit{non-rigid registration} layer to create aligned feature maps. The RNN part is trained with the aligned maps and the rPPG supervision, which examines temporal variability across video frames. 

\begin{figure*}[t!]
	\centering
	\small
	\includegraphics[width=\linewidth]{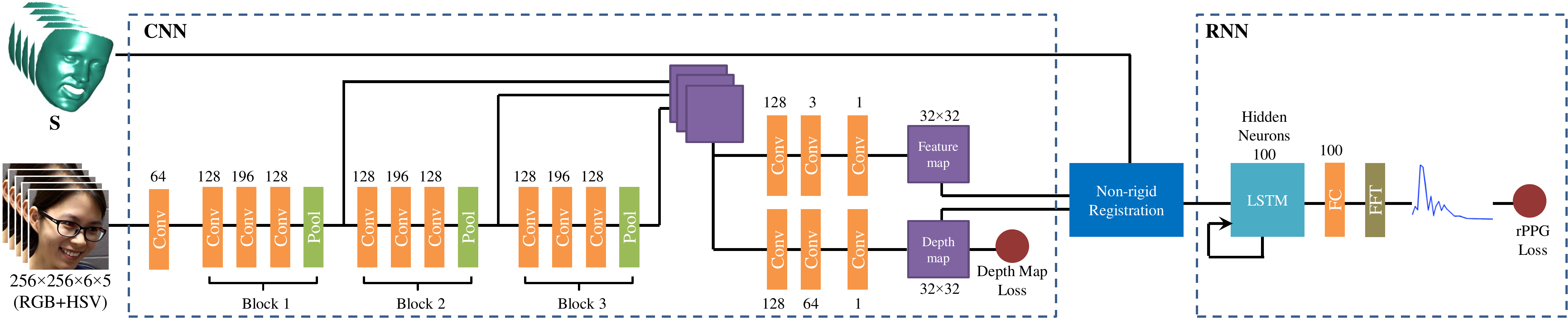} 
	\caption{\small The proposed CNN-RNN architecture. The number of filters are shown on top of each layer, the size of all filters is $3 \times 3$ with stride $1$ for convolutional and $2$ for pooling layers. \textit{Color code} used: \textit{orange}=convolution, \textit{green}=pooling, \textit{purple}=response map.} 
	\label{fig:architecture}
	\figvspace
\end{figure*}

\SubSection{Depth Map Supervision} \label{seq:depthSup}
Depth maps are a representation of the $3$D shape of the face in a $2$D image, which shows the face location and the depth information of different facial areas. 
This representation is more informative than binary labels since it indicates one of the fundamental differences between live faces, and print and replay PA.
We utilize the depth maps in the depth loss function to supervise the CNN part. 
The pixel-based depth loss guides the CNN to learn a mapping from the face area within a receptive field to a labeled depth value -- a scale within $[0,1]$ for live faces and $0$ for spoof faces. 

To estimate the depth map for a $2$D face image, given a face image, we utilize the state-of-the-art dense face alignment (DeFA) methods~\cite{liu2017dense,jourabloo2017pose} to estimate the $3$D shape of the face. 
The frontal dense $3$D shape $\mathbf{S}_F \in \mathbb{R}^{3 \times Q}$, with $Q$ vertices, is represented as a linear combination of identity bases $\{\mathbf{S}_{id}^i\}_{i=1}^{N_{id}}$ and expression bases $\{\mathbf{S}_{exp}^i\}_{i=1}^{N_{exp}}$,   
\eqnvspace
\begin{equation}\label{eq:3DMM}
\mathbf{S}_F=\mathbf{S}_0+\sum_{i=1}^{N_{id}} \alpha_{id}^i \mathbf{S}_{id}^i +\sum_{i=1}^{N_{exp}} \alpha_{exp}^i \mathbf{S}
_{exp}^i,
\end{equation}
where $\mathbf{\alpha}_{id} \in \mathbb{R}^{199}$ and $\mathbf{\alpha}_{ext} \in \mathbb{R}^{29}$ are the identity and expression parameters, and $\mathbf{\alpha}=[\mathbf{\alpha}_{id}, \mathbf{\alpha}_{exp}]$ are the shape parameters. We utilize the Basel $3$D face model~\cite{paysan20093d} and the facewearhouse~\cite{cao2014facewarehouse} as the identity and expression bases.


With the estimated pose parameters $\mathbf{P}=(s, \mathbf{R}, \mathbf{t})$, where $\mathbf{R} $ is a $3$D rotation matrix, $\mathbf{t}$ is a $3$D translation, and $s$ is a scale,  we align the $3$D shape $\mathbf{S}$ to the $2$D 
face image:  
\eqnvspace
\begin{equation}
\mathbf{S}=s\mathbf{R}\mathbf{S}_F + \mathbf{t}.
\end{equation}

Given the challenge of estimating the {\it absolute} depth from a $2$D face, we normalize the $z$ values of $3$D vertices in $\mathbf{S}$ to be within $[0,1]$.
That is, the vertex closest to the camera (e.g., nose) has a depth of one, and the vertex furthest away has the depth of zero.
Then, we apply the Z-Buffer algorithm~\cite{zhu2016face} to $\mathbf{S}$ for projecting the normalized $z$ values to a $2$D plane, which results in an estimated ``ground truth'' $2$D depth map $\mathbf{D} \in \mathbb{R}^{32\times32}$ for a face image.

\SubSection{rPPG Supervision} \label{seq:rppgSup}
rPPG signals have recently been utilized for face anti-spoofing~\cite{liu20163d2,nowara2017ppgsecure}. 
The rPPG signal provides temporal information about face liveness, as it is related to the intensity changes of facial skin over time. 
These intensity changes are highly correlated with the blood flow. 
The traditional method~\cite{de2013robust} for extracting rPPG signals has three drawbacks. 
First, it is sensitive to pose and expression variation, as it becomes harder to {\it track} a specific face area for measuring intensity changes. 
Second, it is also sensitive to illumination changes, since the extra lighting affects the amount of reflected light from the skin. 
Third, for the purpose of anti-spoof, rPPG signals extracted from spoof videos might not be sufficiently {\it distinguishable} to signals of live videos.

One novelty aspect of our approach is that, instead of computing the rPPG signal via~\cite{de2013robust}, our RNN part learns to estimate the rPPG signal. 
This eases the signal estimation from face videos with PIE variations, and also leads to more discriminative rPPG signals, as different rPPG supervisions are provided to live vs.~spoof videos.
We assume that the videos of the same subject under different PIE conditions have the {\it same} ground truth rPPG signal. 
This assumption is valid since the heart beat is similar for the videos of the same subject that are captured in a short span of time ($< 5$ minutes). The rPPG signal extracted from the constrained videos (i.e., no PIE variation) are used as the ``ground truth" supervision in the rPPG loss function for {\it all} live videos of the same subject. 
This consistent supervision helps the CNN and RNN parts to be robust to the PIE changes.


In order to extract the rPPG signal from a face video without PIE, we apply the DeFA~\cite{liu2017dense} to each frame and estimate the dense $3$D face shape. 
We utilize the estimated $3$D shape to track a face region. 
For a tracked region, we compute two orthogonal chrominance signals $\mathbf{x}_f=3\mathbf{r}_f-2\mathbf{g}_f$, $\mathbf{y}_f=1.5\mathbf{r}_f+\mathbf{g}_f-1.5\mathbf{b}_f$ where $\mathbf{r}_f, \mathbf{g}_f, \mathbf{b}_f$ are the bandpass filtered versions of the $\mathbf{r}, \mathbf{g}, \mathbf{b}$ channels with the skin-tone normalization. 
We utilize the ratio of the standard deviation of the chrominance signals $\gamma=\frac{\sigma(\mathbf{x}_f)}{\sigma(\mathbf{y}_f)}$ to compute blood flow signals~\cite{de2013robust}. 
We calculate the  signal $\mathbf{p}$ as:
\eqnvspace
\begin{equation}
\mathbf{p}=3(1-\frac{\gamma}{2})\mathbf{r}_f-2(1+\frac{\gamma}{2})\mathbf{g}_f+\frac{3\gamma}{2}\mathbf{b}_f.
\end{equation}  
By applying FFT to $\mathbf{p}$, we obtain the rPPG signal $\mathbf{f} \in \mathbb{R}^{50}$, which shows the magnitude of each frequency. 

\SubSection{Network Architecture} 
Our proposed network consists of two deep networks. 
First, a CNN part evaluates each frame separately and estimates the depth map and feature map of each frame. 
Second, a recurrent neural network (RNN) part evaluates the temporal variability across the feature maps of a sequence. 

\SubSubSection{CNN Network} 
We design a Fully Convolutional Network (FCN) as our CNN part, as shown in Fig.~\ref{fig:architecture}. 
The CNN part contains multiple blocks of three convolutional layers, pooling and resizing layers where each convolutional layer is followed by one exponential linear layer and batch normalization layer. 
Then, the resizing layers resize the response maps after each block to a pre-defined size of $64\times64$ and concatenate the response maps. 
The bypass connections help the network to utilize extracted features from layers with different depths similar to the ResNet structure~\cite{he2016deep}. 
After that, our CNN has two branches, one for estimating the depth map and the other for estimating the feature map. 

The first output of the CNN is the estimated depth map of the input frame $\mathbf{I} \in \mathbb{R}^{256\times256}$, which is supervised by the estimated ``ground truth" depth $\mathbf{D}$,
\eqnvspace
\begin{equation}\label{eq:depthloss}
\Theta_D = \argmin_{\Theta_D} \sum_{i=1}^{N_d} ||\mbox{CNN}_D( \mathbf{I}_i; \Theta_D) - \mathbf{D}_i ||^2_1,
\end{equation}
where  $\Theta_D$ is the CNN parameters and $N_d$ is the number of training images. 
The second output of the CNN is the feature map, which is fed into the non-rigid registration layer.

\SubSubSection{RNN Network} 
The RNN part aims to estimate the rPPG signal $\mathbf{f}$ of an input sequence with $N_f$ frames $\{\mathbf{I}_j\}_{j=1}^{N_f}$. 
As shown in Fig.~\ref{fig:architecture}, we utilize one LSTM layer with $100$ hidden neurons, one fully connected layer, and an FFT layer that converts the response of fully connected layer into the Fourier domain. 
Given the input sequence $\{\mathbf{I}_j\}_{j=1}^{N_f}$ and the ``ground truth" rPPG signal $\mathbf{f}$, we train the RNN to minimize the $\ell_1$ distance of the estimated rPPG signal to ``ground truth" $\mathbf{f}$,
\eqnvspace
\begin{equation}\label{eq:rppgloss}\small
\Theta_R = \argmin_{\Theta_R} \sum_{i=1}^{N_s} ||\mbox{RNN}_R( [\{\mathbf{F}_j \}_{j=1}^{N_f}]_i; \Theta_R) - \mathbf{f}_i ||^2_1,
\end{equation}
where $\Theta_R$ is the RNN parameters, $\mathbf{F}_j \in \mathbb{R}^{32\times32}$ is the frontalized feature map (details in Sec.~\ref{Sec:Non-rigid_Registration}), and $N_s$ is the number of sequences.

\begin{figure}[t!]
	\centering
	\small
	\includegraphics[width=0.85\linewidth]{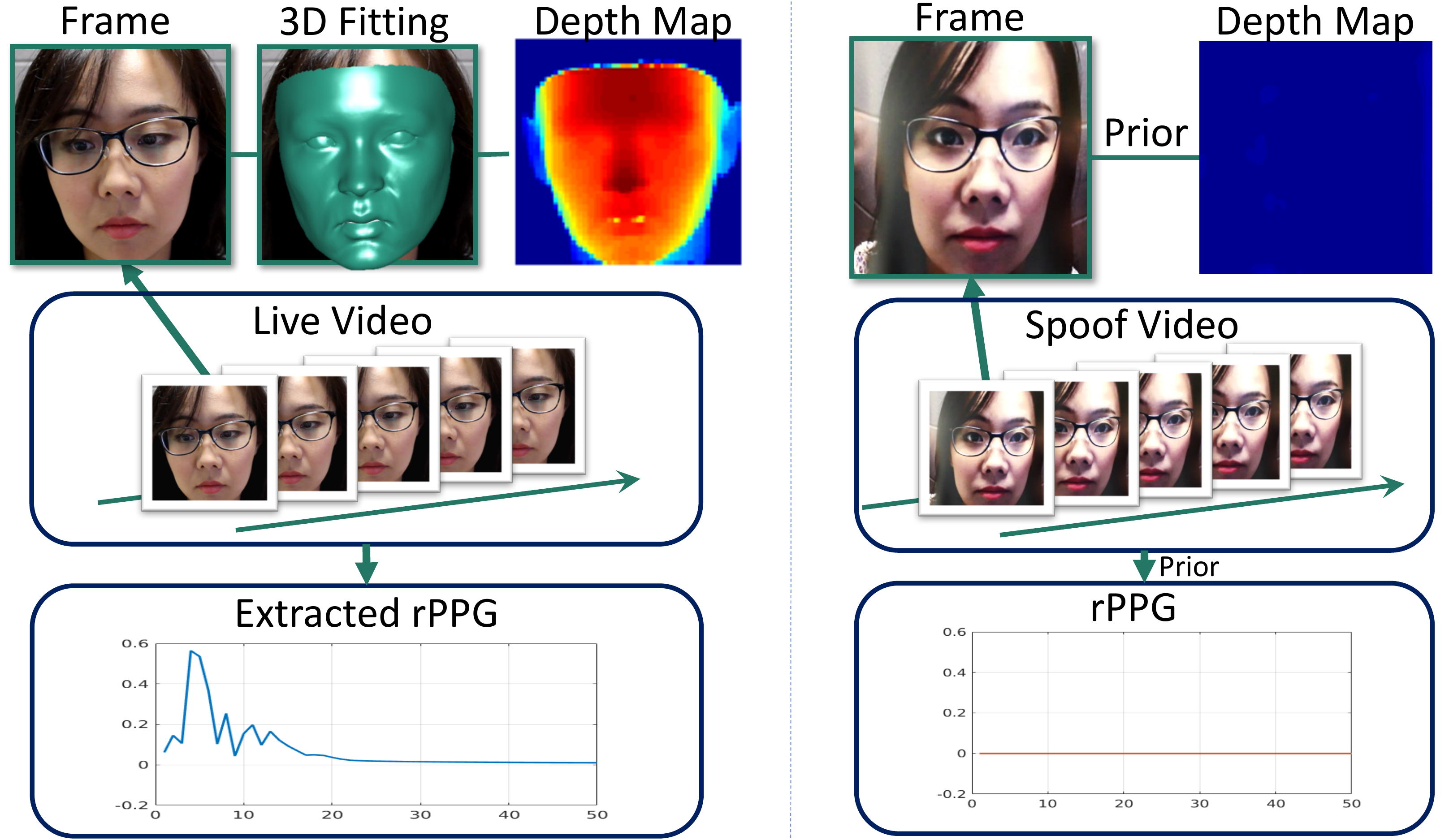} 
	\caption{\small Example ground truth depth maps and rPPG signals.}
	\label{fig:GroundTruth}
	\figvspace
\end{figure}

\SubSubSection{Implementation Details}

\Paragraph{Ground Truth Data}
Given a set of live and spoof face videos, we provide the ground truth supervision for the depth map $\mathbf{D}$ and rPPG signal $\mathbf{f}$, as in Fig.~\ref{fig:GroundTruth}.
We follow the procedure in Sec.~\ref{seq:depthSup} to compute ``ground truth" data for live videos. 
For spoof videos, we set the ground truth depth maps to a plain surface, i.e., zero depth. 
Similarly, we follow the procedure in Sec.~\ref{seq:rppgSup} to compute the ``ground truth" rPPG signal from a patch on the forehead, for one live video of each subject without PIE variation. 
Also, we normalize the norm of estimated rPPG signal such that $\|\mathbf{f}\|_2=1$. 
For spoof videos, we consider the rPPG signals are zero. 

Note that, while the term ``depth" is used here,  our estimated depth is different to the conventional depth map in computer vision.
It can be viewed as a ``pseudo-depth" and serves the purpose of providing discriminative auxiliary supervision to the learning process.
The same perspective applies to the supervision based on pseudo-rPPG signal.

\Paragraph{Training Strategy}
Our proposed network combines the CNN and RNN parts for end-to-end training. 
The desired training data for the CNN part should be from diverse subjects, so as to make the training procedure more stable and increase the generalizability of the learnt model. 
Meanwhile, the training data for the RNN part should be long sequences to leverage the temporal information across frames. 
These two preferences can be contradictory to each other, especially given the limited GPU memory.
Hence, to satisfy both preferences, we design a two-stream training strategy. 
The first stream satisfies the preference of the CNN part, where the input includes face images $\mathbf{I}$ and the ground truth depth maps $\mathbf{D}$. 
The second stream satisfies the RNN part, where the input includes face sequences $\{\mathbf{I}_j \}_{j=1}^{N_f}$, the ground truth depth maps $\{\mathbf{D}_j \}_{j=1}^{N_f}$, the estimated 3D shapes $\{\mathbf{S}_j \}_{j=1}^{N_f}$, and the corresponding ground truth rPPG signals $\mathbf{f}$.  
During training, our method alternates between these two streams to converge to a model that minimizes both the depth map and rPPG losses. 
Note that even though the first stream only updates the weights of the CNN part,  the back propagation of the second stream updates the weights of both CNN and RNN parts in an end-to-end manner. 

\Paragraph{Testing}
To provide a classification score, we feed the testing sequence to our network and compute the depth map $\hat{\mathbf{D}}$ of the last frame and the rPPG signal $\hat{\mathbf{f}}$. 
Instead of designing a classifier using $\hat{\mathbf{D}}$ and $\hat{\mathbf{f}}$, we compute the final score as:
\eqnvspace
\begin{equation}\label{eq:score}
score = ||\hat{\mathbf{f}}||^2_2+\lambda ||\hat{\mathbf{D}}||^2_2,
\end{equation}
where $\lambda$ is a constant weight for combining the two outputs of the network.


\begin{figure}[t!]
	\centering
	\small
	\includegraphics[width=.82\linewidth]{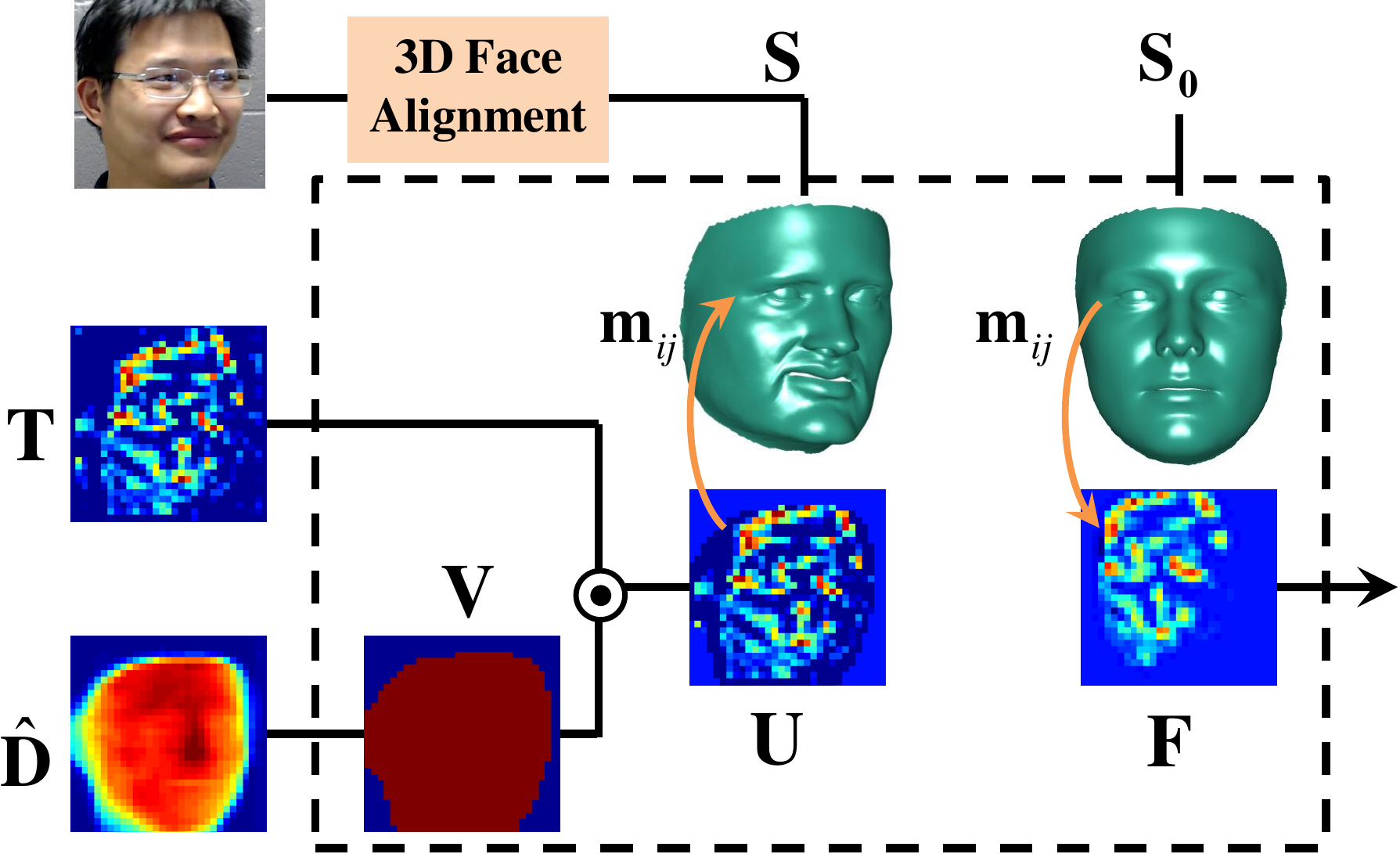} 
	\caption{\small The non-rigid registration layer.}
	\label{fig:dataGen}\figvspace

\end{figure}

\SubSection{Non-rigid Registration Layer} 
\label{Sec:Non-rigid_Registration}
We design a new non-rigid registration layer to prepare data for the RNN part. 
This layer utilizes the estimated dense $3$D shape to align the activations or feature maps from the CNN part.
This layer is important to ensure that the RNN tracks and learns the changes of the activations for the {\it same facial area}  across time, as well as across all subjects.

As shown in Fig.~\ref{fig:dataGen}, this layer has three inputs: the feature map $\mathbf{T} \in \mathbb{R}^{32\times32}$, the depth map $\hat{\mathbf{D}}$ and the $3$D shape $\mathbf{S}$. 
Within this layer, we first threshold the depth map and generate a binary mask $\mathbf{V} \in \mathbb{R}^{32\times32}$: 
\eqnvspace
\begin{equation}\label{eq:depthThresh}
\mathbf{V} = \hat{\mathbf{D}}\ge threshold.
\end{equation}
Then, we compute the inner product of the binary mask and the feature map $\mathbf{U}=\mathbf{T} \odot \mathbf{V}$, which essentially utilizes the depth map as a visibility indicator for each pixel in the feature map. 
If the depth value for one pixel is less than the threshold, we consider that pixel to be invisible. 
Finally, we frontalize $\mathbf{U}$ by utilizing the estimated $3$D shape $\mathbf{S}$, 
\eqnvspace
\begin{equation}\label{eq:Frontal}
\mathbf{F}(i,j) = \mathbf{U}(\mathbf{S}(\mathbf{m}_{ij},1),\mathbf{S}(\mathbf{m}_{ij},2)),
\end{equation}
where $\mathbf{m}\in \mathbb{R}^{K}$ is the pre-defined list of $K$ indexes of the face area in $\mathbf{S_0}$, and $\mathbf{m}_{ij}$ is the corresponding index of pixel $i,j$. 
We utilize $\mathbf{m}$ to project the masked activation map $\mathbf{U}$ to the frontalized image $\mathbf{F}$.   
\begin{table*}[t!]
\small
	\centering
	\caption{The comparison of our collected SiW dataset with existing datasets for face anti-spoofing.}
	\resizebox{\textwidth}{!} 
{
	\begin{tabular}{l|c|c|c|c|c|c|c|c|c}
		\hline
	\multirow{2}{*}{Dataset} &Year& \# of & \# of  & \# of live/attack& Pose & Different & Extra & \multirow{2}{*}{Display devices} & Spoof \\ 
&& subj. & sess. & vid. (V), ima. (I)&range & expres. & light.& & attacks  \\ \hline
		
NUAA~\cite{tan2010face}&$2010$& $15$ & $3$ & $5105/7509 $ (I) & Frontal&No& Yes& -  & Print   \\ \hline		
CASIA-MFSD~\cite{zhang2012face}&$2012$& $50$ & $3$ & $150/450 $ (V) & Frontal& No& No& iPad & Print, Replay   \\ \hline
Replay-Attack~\cite{Chingovska_BIOSIG-2012}&$2012$&$50$& $1$ &$200/1000 $ (V) &  Frontal&No&Yes& iPhone $3$GS, iPad & Print, 2 Replay \\ \hline
MSU-MFSD~\cite{WenTIFS15}&$2015$&$35$& $1$ & $110/330 $ (V) & Frontal &No&No& iPad Air, iPhone $5$S & Print, $2$ Replay \\ \hline
MSU-USSA~\cite{patel2016secure}&$2016$&$1140$& $1$ & $1140/9120 $ (I) &  $[-45^{\circ},45^{\circ}]$&Yes&Yes& MacBook, Nexus $5$, Nvidia Shield Tablet & $2$ print, $6$ Replay \\ \hline

Oulu-NPU~\cite{OULU_NPU_2017}&$2017$&$55$& $3$ & $1980/3960 $ (V) &  Frontal&No&Yes& Dell $1905$FP, Macbook Retina  & $2$ Print, $2$ Replay \\
 \hline

SiW &$2018$&$165$& $4$  & $1320/3300 $ (V) & $[-90^{\circ},90^{\circ}]$&Yes&Yes& iPad Pro, iPhone $7$, Galaxy S$8$, Asus MB$168$B & $2$ Print, $4$ Replay \\ \hline
	       \hline
	\end{tabular}
	} 

\label{tab:dataset}
\figvspace
\end{table*}
This proposed non-rigid registration layer has three  contributions to our network:

$\diamond$ By applying the non-rigid registration, the input data are aligned and the RNN can compare the feature maps without concerning about the facial pose or expression. 
In other words, it can learn the temporal changes in the activations of the feature maps for the same facial area.

$\diamond$ The non-rigid registration removes the background area in the feature map. Hence the background area would not participate in RNN learning, although the background information is already utilized in the layers of the CNN part.

$\diamond$ For spoof faces, the depth maps are likely to be closer to zero. Hence, the inner product with the depth maps substantially weakens the activations in the feature maps, which makes it easier for the RNN to output zero rPPG signals. Likewise, the back propagation from the rPPG loss also encourages the CNN part to generate zero depth maps for either all frames, or one pixel location in majority of the frames within an input sequence.


	
%

\begin{figure}[t!]
	\centering
	\small
		\includegraphics[height=.5\linewidth]{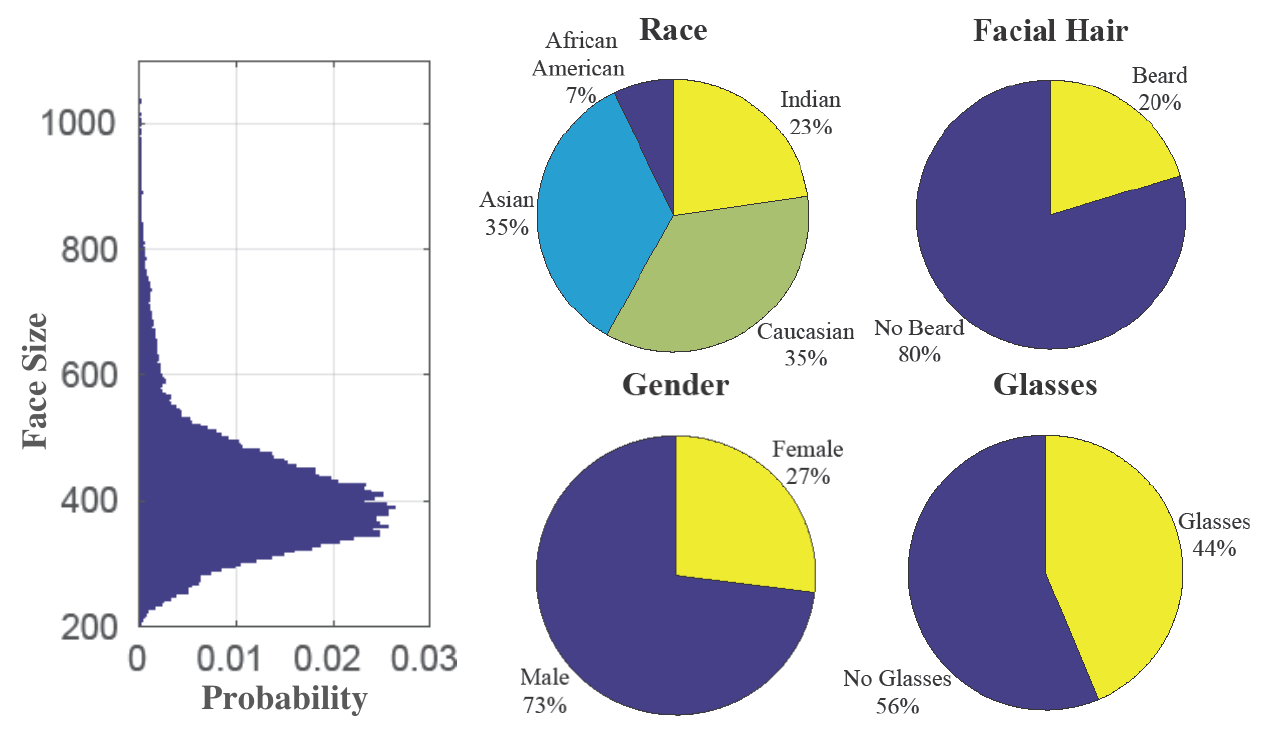} 
	\caption{\small The statistics of the subjects in the SiW database. Left side: The histogram shows the distribution of the face sizes.}
	\label{fig:stat}
	\figvspace
\end{figure}


\Section{Collection of Face Anti-Spoofing Database} \label{sec:database}

With the advance of sensor technology, existing anti-spoofing systems can be vulnerable to emerging high-quality spoof mediums. 
One way to make the system robust to these attacks is to collect new high-quality databases. 
In response to this need, we collect a new face anti-spoofing database named Spoof in the Wild (SiW) database, which has multiple advantages over previous datasets as in Tab.~\ref{tab:dataset}. 
First, it contains substantially more live subjects with diverse races, e.g., $3$ times of the subjects of Oulu-NPU.
Note that MSU-USSA is constructed using existing images of celebrities without capturing live faces.
Second, live videos are captured with two high-quality cameras (Canon EOS T$6$, Logitech C$920$ webcam) with different PIE variations.


SiW provides live and spoof $30$-fps videos from $165$ subjects. 
For each subject, we have $8$ live and $20$ spoof videos, in total $4,620$ videos. 
Some statistics of the subjects are shown in Fig.~\ref{fig:stat}. 
The live videos are collected in four sessions. 
In Session $1$, the subject moves his head with varying distances to the camera.
In Session $2$, the subject changes the yaw angle of the head within $[-90^{\circ},90^{\circ}]$, and makes different face expressions.
In Sessions $3,4$, the subject repeats the Sessions $1,2$, while the collector moves the point light source around the face from different orientations. 

\begin{figure}[t!]
	\centering
	\small
	\includegraphics[width=1\linewidth]{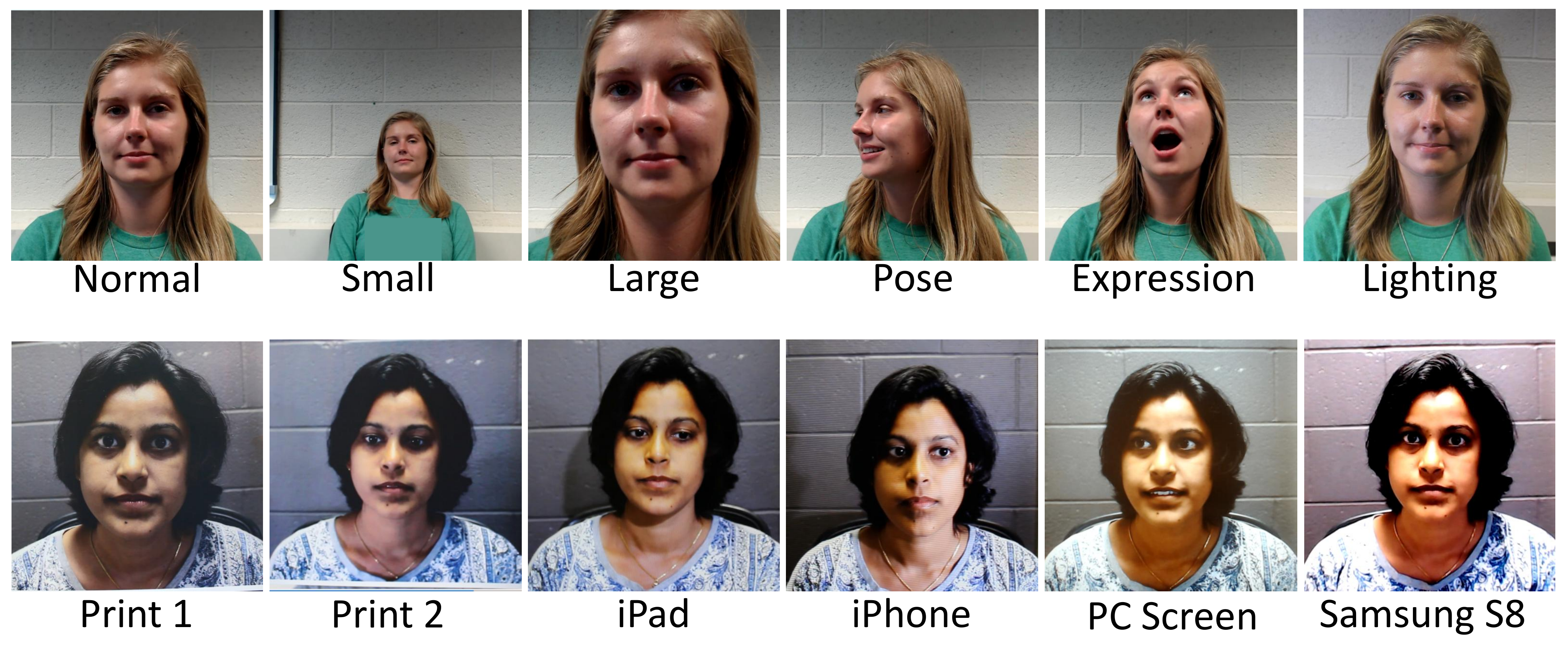} 
\vspace{-5mm}
	\caption{\small Example live (top) and spoof (bottom) videos in SiW.} 
	\label{fig:Siw}
		\figvspace

\end{figure}

The live videos captured by both cameras are of $1,920\times1,080$ resolution.  
We provide two print and four replay video attacks for each subject, with examples shown in Fig.~\ref{fig:Siw}. 
To generate different qualities of print attacks,
we capture a high-resolution image ($5,184\times3,456$) for each subject and use it to make a high-quality print attack. 
Also, we extract a frontal-view frame from a live video for lower-quality print attack. 
We print the images with an HP color LaserJet M652 printer. 
The print attack videos are captured by holding printed papers still or warping them in front of the cameras.
To generate high-quality replay attack videos, we select four spoof mediums: Samsung Galaxy S$8$, iPhone $7$, iPad Pro, and PC (Asus MB$168$B) screens. 
For each subject, we randomly select two of the four high-quality live videos to display in the spoof mediums.

\Section{Experimental Results}

\SubSection{Experimental Setup}
\Paragraph{Databases} We evaluate our method on multiple databases to demonstrate its generalizability. 
We utilize SiW and Oulu databases~\cite{OULU_NPU_2017} as new high-resolution databases and perform intra and cross testing between them. Also, we use the CASIA-MFSD~\cite{zhang2012face} and Replay-Attack~\cite{Chingovska_BIOSIG-2012} databases for cross testing and comparing with the state of the art. 

\Paragraph{Parameter setting} 
The proposed method is implemented in TensorFlow~\cite{tensorflow2015-whitepaper} with a constant learning rate of $3\mathrm{e}{-3}$, and $10$ epochs of the training phase. 
The batch size of the CNN stream is $10$ and that of the CNN-RNN stream is $2$ with $N_f$ being $5$. 
We randomly initialize our network by using a normal distribution with zero mean and std of $0.02$.
We set $\lambda$ in Eq.~\ref{eq:score} to $0.015$ and $threshold$ in Eq.~\ref{eq:depthThresh}  to $0.1$.

\Paragraph{Evaluation metrics} 
To compare with prior works, we report our results with the following metrics: Attack Presentation Classification Error Rate $APCER$~\cite{acer1}, Bona Fide Presentation Classification Error Rate $BPCER$~\cite{acer1}, $ACER=\frac{APCER+BPCER}{2}$~\cite{acer1}, and Half Total Error Rate $HTER$. 
The $HTER$ is half of the summation of the False Rejection Rate (FRR) and the False Acceptance Rate (FAR).

\SubSection{Experimental Comparison}

\begin{table}[t!]
\small
	\centering
	\caption{TDR at different FDRs, cross testing on Oulu Protocol $1$.}
{
	\begin{tabular}{|c|c|c|c|c|}
		\hline
FDR & $1\%$ & $2\%$ & $10\%$ & $20\%$\\ \hline
Model $1$ & $8.5\%$ & $18.1\%$ & $71.4\%$ & $81.0\%$\\ \hline
Model $2$ & $40.2\%$ & $46.9\%$ & $78.5\%$ & $93.5\%$\\ \hline
Model $3$ & $39.4\%$ & $42.9\%$ & $67.5\%$ & $87.5\%$\\ \hline
Model $4$ & $\textbf{45.8\%}$ & $\textbf{47.9\%}$ & $\textbf{81\%}$ & $\textbf{94.2\%}$\\ \hline

	\end{tabular}
	} 

\label{tab:abolation}
\figvspace
\end{table}

\begin{table}[t!]
\small
	\centering
	\caption{\small ACER of our method at different $N_f$, on Oulu Protocol $2$.}

\scalebox{1}{
	\begin{tabular}{|c|c|c|c|}
		\hline
\backslashbox{Test}{Train}& $5$ & $10$ & $20$\\ \hline		
$5$ & $4.16\%$ & $4.16\%$ & $3.05\%$ \\ \hline
$10$ & $4.02\%$ & $3.61\%$ & $2.78\%$ \\ \hline
$20$ & $4.10\%$ & $3.67\%$ & $2.98\%$ \\ \hline

	\end{tabular}
	} 

\label{tab:abolation2}
\figvspace
\end{table}

\SubSubSection{Ablation Study}

\Paragraph{Advantage of proposed architecture} 
We compare four architectures to demonstrate the advantages of the proposed loss layers and non-rigid registration layer.
\textit{Model} $1$ has an architecture similar to the CNN part in our method (Fig.~\ref{fig:architecture}), except that it is extended with additional pooling layers, fully connected layers, and softmax loss for binary classification.
\textit{Model} $2$ is the CNN part in our method with a depth map loss function. We simply use $||\mathbf{\hat{D}}||_2$ for classification.
\textit{Model} $3$ contains the CNN and RNN parts without the non-rigid registration layer. 
Both of the depth map and rPPG loss functions are utilized in this model.
However, the RNN part would process unregistered feature maps from the CNN.
\textit{Model} $4$ is the proposed architecture.

We train all four models with the live and spoof videos from $20$ subjects of SiW. 
We compute the cross-testing performance of all models on Protocol $1$ of Oulu database. 
The TDR at different FDR are reported in Tab.~\ref{tab:abolation}. 
\textit{Model} $1$ has a poor performance due to the binary supervision. 
In comparison, by only using the depth map as supervision, \textit{Model} $2$ achieves substantially better performance.
However, after adding the RNN part with the rPPG supervision, our proposed \textit{Model} $4$ can further the performance improvement.
By comparing \textit{Model} $4$ and $3$, we can see the advantage of the non-rigid registration layer. 
It is clear that the RNN part cannot use feature maps directly for tracking the changes in the activations and estimating the rPPG signals. 

\Paragraph{Advantage of longer sequences} 
To show the advantage of utilizing longer sequences for estimating the rPPG, we train and test our model when the sequence length $N_f$ is $5$, $10$, or $20$, using intra-testing on Oulu Protocol $2$. 
From Tab.~\ref{tab:abolation2}, we can see that by increasing the sequence length, the ACER decreases due to more reliable rPPG estimation. 
Despite the benefit of longer sequences, in practice, we are limited by the GPU memory size, and forced to decrease the image size to $128 \times 128$ for all experiments in Tab.~\ref{tab:abolation2}. 
Hence, we set $N_f$ to be $5$ with the image size of $256\times256$ in subsequent experiments, due to importance of higher resolution (e.g, a lower \textit{ACER} of $2.5\%$ in Tab.~\ref{tab:intraOulu} is achieved than $4.16\%$).

\SubSubSection{Intra Testing} We perform intra testing on Oulu and SiW databases. 
For Oulu, we follow the four protocols~\cite{boulkenafet17ijcb} and report their \textit{APCER}, \textit{BPCER} and \textit{ACER}.
Tab.~\ref{tab:intraOulu} shows the comparison of our proposed method and the best two methods for {\it each} protocol respectively, in the face anti-spoofing competition~\cite{boulkenafet17ijcb}. 
Our method achieves the lowest \textit{ACER} in $3$ out of $4$ protocols. 
We have slightly worse \textit{ACER} on Protocol $2$.
To set a baseline for future study on SiW, we define three protocols for SiW. 
The Protocol $1$ deals with variations in face pose and expression. 
We train using the first $60$ frames of the training videos that are mainly frontal view faces,  and test on all testing videos. 
The Protocol $2$ evaluates the performance of cross spoof medium of replay attack. 
The Protocol $3$ evaluates the performance of cross PA, i.e., from print attack to replay attack and vice versa. 
Tab.~\ref{tab:ProtSiW} shows the protocol definition and our performance of each protocol.

\begin{table}[t!]
\small
	\centering
	\caption{The intra-testing results on four protocols of Oulu.}
	\resizebox{0.48\textwidth}{!} 
{
	\begin{tabular}{|c|c|c|c|c|}
		\hline
Prot. &Method & APCER (\%) & BPCER (\%) & ACER (\%)\\ \hline  \hline
    & CPqD & $2.9$ & $10.8$ & $6.9$\\ \cline{2-5}
$1$ & GRADIANT & $\textbf{1.3}$ & $12.5$ & $6.9$\\ \cline{2-5}
   & Proposed method & $1.6$ & $\textbf{1.6}$ & $\textbf{1.6}$\\ \hline \hline
   & MixedFASNet & $9.7$ & $2.5$ & $6.1$\\ \cline{2-5}
$2$ & Proposed method & $\textbf{2.7}$ & $2.7$ & $2.7$\\ \cline{2-5}
  & GRADIANT & $3.1$ & $\textbf{1.9}$ & $\textbf{2.5}$\\ \hline \hline
 &MixedFASNet & $5.3\pm6.7$ & $7.8\pm5.5$ & $6.5\pm4.6$\\  \cline{2-5}
$3$ &GRADIANT & $\textbf{2.6}\pm\textbf{3.9}$ & $5.0\pm5.3$ & $3.8\pm2.4$\\  \cline{2-5}
 & Proposed method & $2.7\pm1.3$ & $\textbf{3.1}\pm\textbf{1.7}$ & $\textbf{2.9}\pm\textbf{1.5}$\\ \hline \hline
  & Massy\_HNU & $35.8\pm35.3$ & $\textbf{8.3}\pm\textbf{4.1}$ & $22.1\pm17.6$\\ \cline{2-5}
$4$ & GRADIANT & $\textbf{5.0}\pm\textbf{4.5}$ & $15.0\pm7.1$ & $10.0\pm5.0$\\ \cline{2-5}
  & Proposed method & $9.3\pm5.6$ & $10.4\pm6.0$ & $\textbf{9.5}\pm\textbf{6.0}$\\ \hline

	\end{tabular}
	} 

\label{tab:intraOulu}
\figvspace
\end{table}

\begin{table}[t!]
\small
	\centering
	\caption{The intra-testing results on three protocols of SiW.}
	\resizebox{0.48\textwidth}{!} 
{
	\begin{tabular}{|c|c|c|c|c|c|c|}
		\hline
Prot. & Subset & Subject \# & Attack & APCER (\%) & BPCER (\%) & ACER (\%)\\ \hline
\multirow{2}{*}{$1$} & Train & $90$ & First $60$ Frames  & \multirow{2}{*}{$3.58$} & \multirow{2}{*}{$3.58$} & \multirow{2}{*}{$3.58$}\\ \cline{2-4}
&Test & $75$ & All & & &\\\hline
\multirow{2}{*}{$2$} & Train& $90$ & $3$ display &\multirow{2}{*}{$0.57 \pm 0.69$} & \multirow{2}{*}{$0.57 \pm 0.69$}  & \multirow{2}{*}{$0.57 \pm 0.69$}\\  \cline{2-4}
&Test &$75$ & $1$ display & & &\\\hline
\multirow{2}{*}{$3$} & Train&$90$& print (display) &\multirow{2}{*}{$8.31 \pm 3.81$} & \multirow{2}{*}{$8.31 \pm 3.80$} & \multirow{2}{*}{$8.31 \pm 3.81$}\\  \cline{2-4}
&Test &$75$ & display (print) & & &\\\hline

	\end{tabular}
	} 

\label{tab:ProtSiW}
\figvspace
\end{table}

\begin{figure*}[t!]
	\centering
	\small
\includegraphics[width=\linewidth]{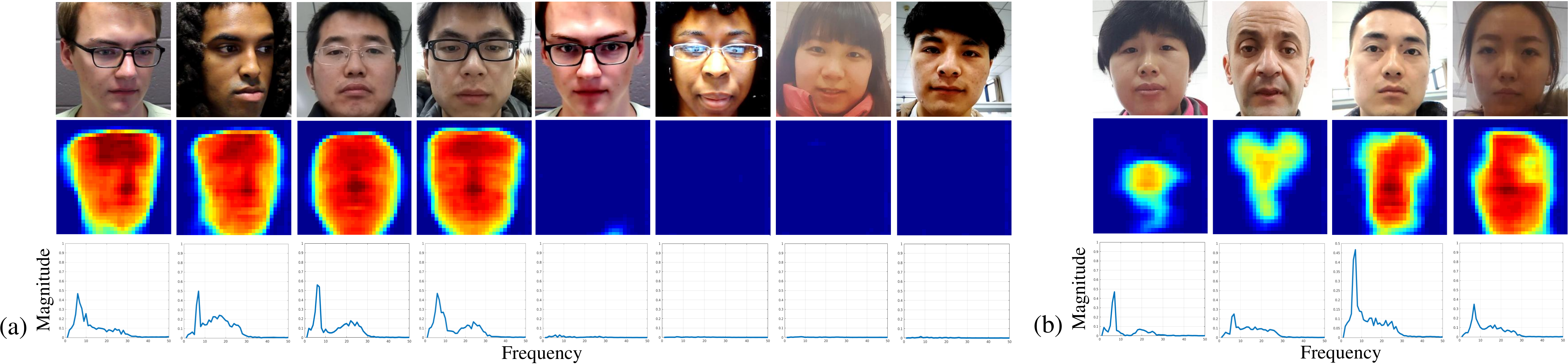}	\caption{\small (a) $8$ successful anti-spoofing examples and their estimated depth maps and rPPG signals. (b) $4$ failure examples: the first two  are live and the other two are spoof. Note our ability to estimate discriminative depth maps and rPPG signals.}
	\label{fig:example}
	\figvspace
\end{figure*}

\SubSubSection{Cross Testing}
To demonstrate the generalization of our method, we perform multiple cross-testing experiments. 
Our model is trained with live and spoof videos of $80$ subjects in SiW, and test on all protocols of Oulu. 
The \textit{ACER} on Protocol $1$-$4$ are respectively: $10.0\%$, $14.1\%$, $13.8\pm5.7\%$, and $10.0\pm8.8\%$. 
Comparing these cross-testing results to the {\it intra-testing} results in~\cite{boulkenafet17ijcb}, we are ranked sixth on the average \textit{ACER} of four protocols, among the $15$ participants of the face anti-spoofing competition.
Especially on Protocol $4$, the hardest one among all protocols, we achieve the {\it same} \textit{ACER} of $10.0\%$ as the top performer.
This is a notable result since cross testing is known to be substantially harder than intra testing, and yet our cross-testing result is comparable with the top intra-testing performance.
This demonstrates the generalization ability of our learnt model.

Furthermore, we utilize the CASIA-MFSD and Replay-Attack databases to perform cross testing between them, which is widely used as a cross-testing benchmark.
Tab.~\ref{tab:cross} compares the cross-testing \textit{HTER} of different methods. 
Our proposed method reduces the cross-testing errors on the Replay-Attack and CASIA-MFSD databases by $8.9\%$ and $24.6\%$ respectively, relative to the previous SOTA.

\begin{table}[t!]
\tiny
	\centering
	\caption{ \small Cross testing on CASIA-MFSD vs.~Replay-Attack.}
	\resizebox{\linewidth}{!} {
	\begin{tabular}{|c|c|c|c|c|}
	
		\hline
\multirow{3}{*}{Method} & Train  & Test& Train & Test \\  \cline{2-5}
& CASIA- & Replay& Replay & CASIA- \\
& MFSD & Attack& Attack & MFSD \\ \hline
Motion~\cite{de2013can} & \multicolumn{2}{c}{50.2\%} & \multicolumn{2}{|c|}{47.9\%} \\  \hline
LBP~\cite{de2013can} & \multicolumn{2}{c}{55.9\%} & \multicolumn{2}{|c|}{57.6\%} \\  \hline
LBP-TOP~\cite{de2013can} & \multicolumn{2}{c}{49.7\%} & \multicolumn{2}{|c|}{60.6\%} \\  \hline
Motion-Mag~\cite{bharadwaj2013computationally} & \multicolumn{2}{c}{50.1\%} & \multicolumn{2}{|c|}{47.0\%} \\  \hline
Spectral cubes~\cite{pinto2015face} & \multicolumn{2}{c}{34.4\%} & \multicolumn{2}{|c|}{50.0\%} \\  \hline
CNN~\cite{yang2014learn} & \multicolumn{2}{c}{48.5\%} & \multicolumn{2}{|c|}{45.5\%} \\  \hline
LBP~\cite{boulkenafet2015face} & \multicolumn{2}{c}{47.0\%} & \multicolumn{2}{|c|}{39.6\%} \\  \hline
Colour Texture~\cite{boulkenafet2016face} & \multicolumn{2}{c}{30.3\%} & \multicolumn{2}{|c|}{37.7\%} \\  \hline
Proposed method & \multicolumn{2}{c}{\textbf{27.6\%}} & \multicolumn{2}{|c|}{\textbf{28.4\%}} \\  \hline

	\end{tabular}
	} 

\label{tab:cross}
\figvspace
\end{table}

\begin{figure}[t!]
	\vspace{1mm}
	\centering
	\small
	\includegraphics[width=\linewidth]{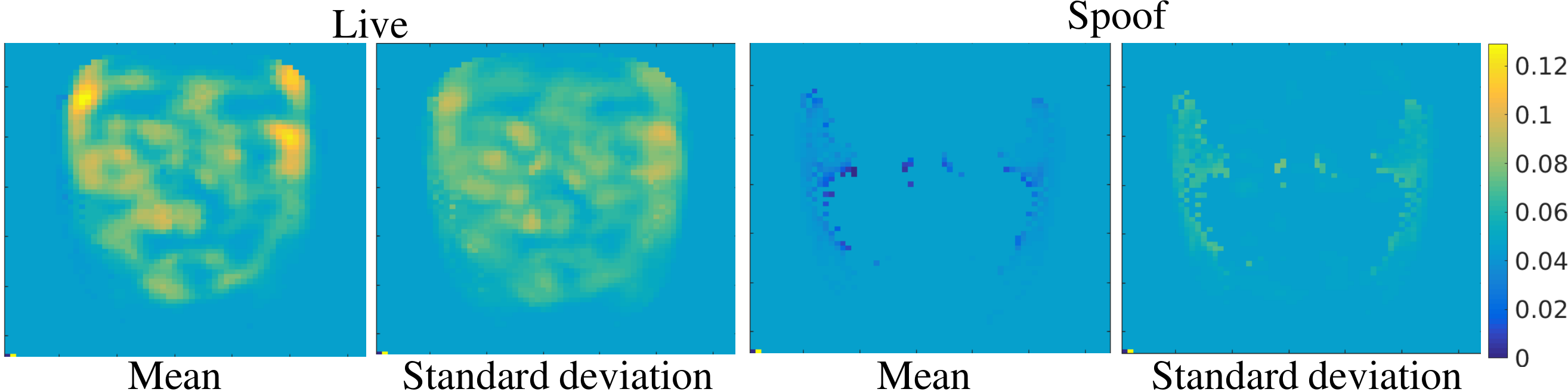} 
	\caption{\small Mean/Std of frontalized feature maps for live and spoof.}
	\label{fig:analysis}
	\figvspace
\end{figure}

\figvspace
\SubSubSection{Visualization and Analysis}
In the proposed architecture, the frontalized feature maps are utilized as input to the RNN part and are supervised by the rPPG loss function. 
The values of these maps can show the importance of different facial areas to rPPG estimation. 
Fig.~\ref{fig:analysis} shows the mean and standard deviation of frontalized feature maps, computed from $1,080$ live and spoof videos of Oulu. 
We can see that the side areas of forehead and cheek have higher influence for rPPG estimation.

\begin{figure}[t!]
	\centering
	\small
	\includegraphics[width=\linewidth]{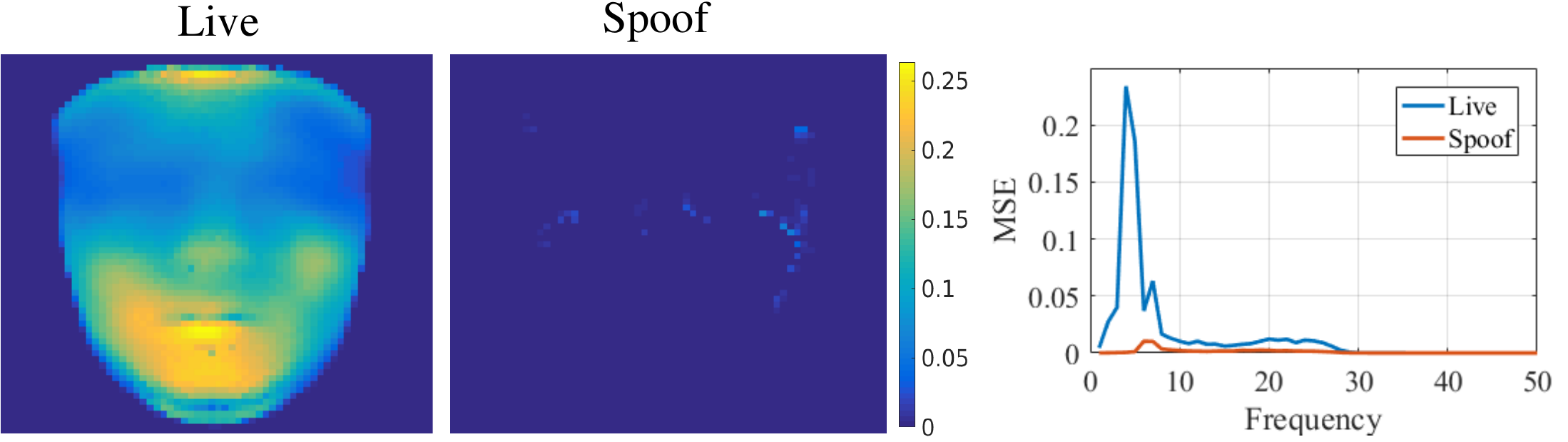} 
	\caption{\small The MSE of estimating depth maps and rPPG signals.}
	\label{fig:analysis2}
	\figvspace
\end{figure}

While the goal of our system is to detect PAs, our model is trained to estimate the auxiliary information.
Hence, in addition to anti-spoof, we also like to evaluate the accuracy of auxiliary information estimation.
For this purpose, we calculate the accuracy of estimating depth maps and rPPG signals, for testing data in Protocol $2$ of Oulu.
As shown in Fig.~\ref{fig:analysis2}, the accuracy for both estimation in spoof data is high, while that of the live data is relatively lower.
Note that the depth estimation of the mouth area has more errors, which is consistent with the fewer activations of the same area in Fig.~\ref{fig:analysis}.
Examples of successful and failure cases in estimating depth maps and rPPG signals are shown in Fig.~\ref{fig:example}.

Finally, we conduct statistical analysis on the failure cases, since our system can determine potential causes using the auxiliary information.
With Proctocol $2$ of Oulu, we identify $31$ failure cases ($2.7\%$ \textit{ACER}).
For each case, we calculate whether anti-spoofing using its depth map or rPPG signal would fail if that information alone is used.
In total, $\frac{29}{31}$, $\frac{13}{31}$, and $\frac{11}{31}$ samples fail due to depth map, rPPG signals, or both.
This indicates the future research direction.

\Section{Conclusions}
This paper identifies the importance of auxiliary supervision to deep model-based face anti-spoofing. 
The proposed network combines CNN and RNN architectures to jointly estimate the depth of face images and rPPG signal of face video. 
We introduce the SiW database that contains more subjects and variations than prior databases. 
Finally, we experimentally demonstrate the superiority of our method.

\textvspace

\paragraph{Acknowledgment}
This research is based upon work supported by the Office of the Director of National Intelligence (ODNI), Intelligence Advanced Research Projects Activity (IARPA), via IARPA R\&D Contract No.~$2017$-$17020200004$. The views and conclusions contained herein are those of the authors and should not be interpreted as necessarily representing the official policies or endorsements, either expressed or implied, of the ODNI, IARPA, or the U.S. Government. The U.S. Government is authorized to reproduce and distribute reprints for Governmental purposes notwithstanding any copyright annotation thereon.

{\small
\bibliographystyle{ieee}
\bibliography{abbrev_brief,egbib}
}

\end{document}